\documentclass[10pt,letterpaper]{article}
\usepackage[top=0.85in,left=2.75in,footskip=0.75in,marginparwidth=2in]{geometry}

\usepackage[utf8]{inputenc}

\usepackage{cite}

\usepackage{nameref,hyperref}

\usepackage{microtype}
\DisableLigatures[f]{encoding = *, family = * }

\raggedright
\usepackage{changepage}

\usepackage[aboveskip=1pt,labelfont=bf,labelsep=period,singlelinecheck=off]{caption}

\makeatletter
\renewcommand{\@biblabel}[1]{\quad#1.}
\makeatother

\usepackage{lastpage,fancyhdr,graphicx}
\usepackage{epstopdf}
\fancyheadoffset[L]{2.25in}
\fancyfootoffset[L]{2.25in}

\usepackage{color}

\definecolor{Gray}{gray}{.25}

\usepackage{graphicx}

\usepackage{sidecap}

\usepackage{wrapfig}
\usepackage[pscoord]{eso-pic}
\reversemarginpar

\begin{document}
\vspace*{0.35in}

\begin{flushleft}
{\Large
\textbf\newline{Deployment of the Saddle Space Transformation in Tracking the Base of Support}
}
\newline
\\
Carlo Tiseo\textsuperscript{1,2,*},
Ming Jeat Foo\textsuperscript{1,2},
Kalyana C Veluvolu \textsuperscript{3},
Wei Tech Ang\textsuperscript{1,2},
\\
\bigskip
\bf{1} Rehabilitation Research Institute of Singapore, 50 Nanyang Avenue N3-01a-01,Singapore 639798, Singapore;
\\
\bf{2} School of Mechanical \& Aerospace Engineering, Nanyang Technological University, 50 Nanyang Avenue N3-01a-01,Singapore 639798, Singapore;
\\
\bf{3} School of Electronics Engineering,Kyungpook National University, Daegu  702701, South Korea;
\\
\bigskip
* carlotiseo@gmail.com

\end{flushleft}

\section*{Abstract}
Balance is the fundamental skill behind human locomotion, and its impairment is the principal indicator of self-perceived disability. Despite significant improvements in balance assessment, there is still large incidence of fall related injuries among elderlies. The Base of Support (BoS) is a popular method for bipedal stability assessment, but its accuracy depends on the accuracy the BoS geometry measurement. This work presents a method to ease the BoS tracking by the identification of a reference frame that allows to define postural models of the BoS geometry. Although we also propose a geometry based on the geometry determined from centre of pressure range of motion within the foot obtained from literature, this methodology can be used with other models (i.e., the feasible base of support). The model has been tested with 12 healthy subjects, which have been asked to explore their stability in six different postures. The results show that the model can accurate deform the geometry of the BoS to adapt its shape to the different postures, which can remove the necessity of force/torque sensors in some application. Potentially the proposed method can be also applied to describe any posture dependent attribute (e.g., gravitational forces), and it can be also applied to bipedal robots. Therefore, it constitutes a novel mathematical tool that can be deployed to develop both better sensors and models for bipeds. For example, it can be used with the Extrapolated CoM model to evaluate dynamic stability from the body kinematics.

\section*{Introduction}
Human bipedal balance is a fundamental skill that is required for the execution of standing activities \cite{Huxham2001,Pollock2000}. Despite the challenging dynamics involved in the balance mechanism, humans are peculiarly effective in controlling their equilibrium in unstructured environments \cite{Rosso2013}. Furthermore, the human stability control is affected by the physiological ageing due to the increase of reaction times to perturbation \cite{Huxham2001,Maki2007,Rosso2013}. This often leads to mobility impairments which significantly affects the independence and safety of the elderly population \cite{Rosso2013}. Despite the relevance to rapidly ageing societies, there has not been any substantial improvement of the rehabilitation therapies in the last decades \cite{Geurts2005,Diaz2011,Torricelli2016,Maggioni2016,Winstein2016}. 

The balance as defined by Pollock et al \cite{Pollock2000} involves three activities: the maintenance of the desired posture, the voluntary change between postures, and the reaction to external perturbations. Currently, the balance assessment is mainly based on the analysis Base of Support (BoS) and Centre of Gravity (or Centre of Mass, CoM),where posture and movements are defined stable as long as the CoM position is contained within the BoS \cite{Pollock2000, VanMeulen2016}. The accuracy of such approach depends an accurate tracking of both the BoS and CoM, which can be challenging in unstructured environments being both of them posture dependant quantities \cite{VanMeulen2016}. Nonetheless, this approach has been widely adopted in human studies due to its good accuracy and low computational cost, there is not standard method to track BoS geometry \cite{Hof2005,Lugade2011,McAndrewYoung2012,Torricelli2016,Hak2013,Hak2013a,Font-Llagunes2009}. Usually, the BoS is identified as the smallest convex hull enclosing the contact the area of the feet with the ground, which includes the area between them during the double support phase \cite{Yang2009}. 

Recently, our group used the potential energy surface produced by the two legs to analyse human balance during walking. We defined a posture dependant space, called Saddle Space, to study the potential energy of the system during posture changes (Figure \ref{fig:01}.(a)). The main advantage of this space is the possibility to define a single model for BoS geometry that can be projected in the task space \cite{Tiseo2016, Tiseo2018, Tiseo2018a, Tiseo2018b}. The Saddle Space model is based on the identification of a reference point within the foot (i.e., extrapolated Centre of Pressure or eCoP), and a Base of Support (BoS) that describes the range of motion of the physical the centre of pressure within the foot. The position of the eCoP and the shape BoS have been defined according to the data reported by Hof et al \cite{Hof2005} as described in \ref{Apx:1}. Furthermore, we have also conducted preliminary validation to test the hypothesis that the Saddle Space transformation can accurately track BoS geometry comparing the model output with experimental results reported in literature \cite{Tiseo2016}. Lastly, we have also proved in \cite{Tiseo2018} that tracking the BoS evolution during locomotion is a generalization of the extrapolated CoM model proposed by Hof et al \cite{Hof2005}. In fact, the extrapolated CoM provides the inclination of $y_{Saddle}$ for a step frequency equal to the natural frequency of the inverted pendulum,  which describes the minimum step required to achieve stability at a selected velocity \cite{Tiseo2018, Tiseo2018a, Tiseo2018b,Tiseo2018bioinspired}.

This manuscript develops a posture dependent model to track the BoS based on the foot kinematics, which can now be deployed for the development of sensor to track balance assessment and supervision in daily living environments. To validate the accuracy of the proposed model, 12 subjects have been asked to explore their stability in the six different postures as represented in Figure \ref{fig:01}.(b). The results show that the transformation of the Saddle Space is an efficient methodology for the analysis of the bipeds, even though the proposed CoM geometrical model does not fully model the region where healthy subjects can stabilize their body. This reduced accuracy was expected because the proposed model consider the CoM as a fixed within the human body while in reality CoM is posture dependent. Nonetheless, given the difficulty in tracking the human posture and inertias accurately, the geometrical model is usually considered a good approximation \cite{Pai1997,Hof2005,VanMeulen2016,Lugade2011}.

The BoS model used in this paper and the experimental set-up are described in Section \ref{Sec:2}. The results are presented in Section \ref{Sec:3} and discussed in Section \ref{Sec:4}, which also analyses the BoS model limitations and possible improvements by using the Capture Regions (CR) used for bipedal robots \cite{Pratt2006}. Section \ref{Sec:5} draws the conclusions.

\section*{Materials and Methods}

\label{Sec:2}
This Section starts with the presentation of the model employed for the BoS and, subsequently, it describes the experimental method for validation.
\subsection*{Base of Support Model}
\begin{figure}[htbp]
	\begin{center}
		
		\includegraphics[width=\textwidth]{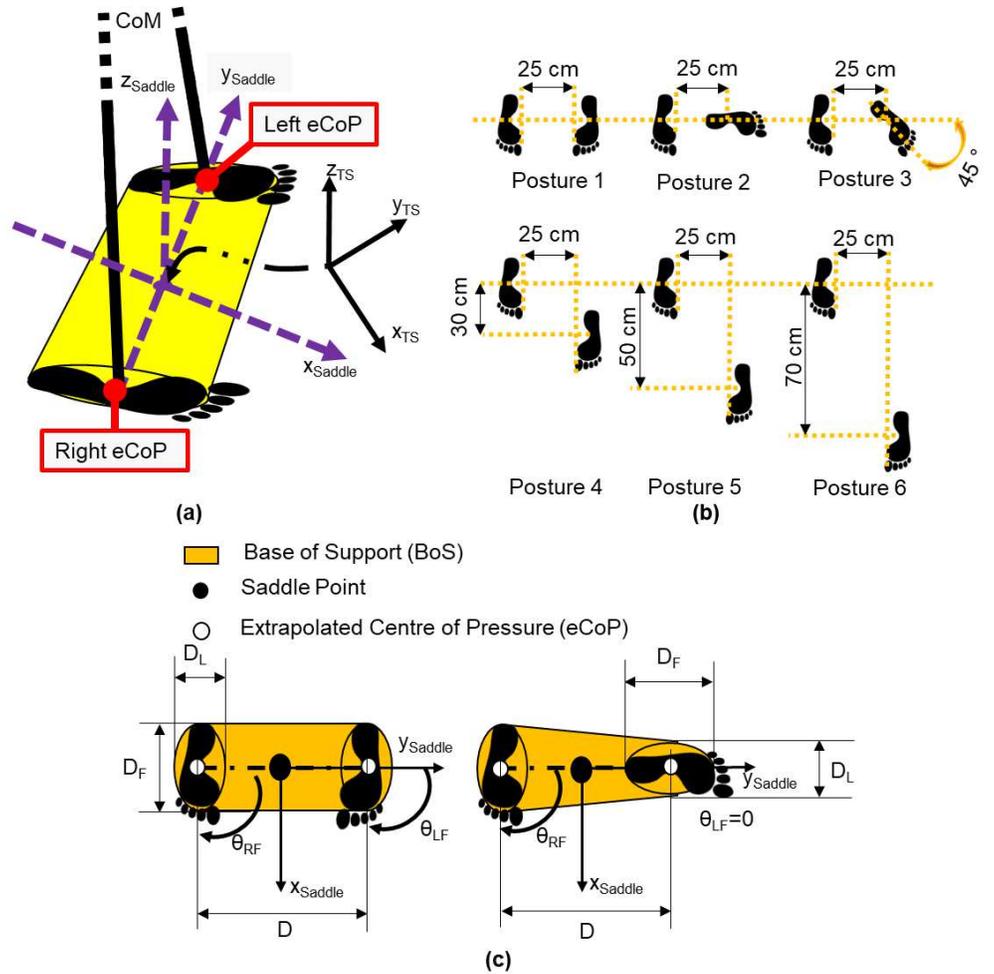}
		
	\end{center}
	\caption{(a) The Saddle frame is a reference system oriented with the principal direction of the potential energy surface generated by the two legs, and it depends from the foot placement. Therefore, it can be a useful tool for describing posture dependent quantities and geometries like the Base of Support  (BoS). (b) Six postures are analysed in this work to validate the hypothesis that the BoS geometry defined in the Saddle Space can be projected in the task space using Equation \ref{eq:01}. (c) The BoS and the Saddle frame of the system when the feet are parallel and orthogonal to each other, graphically represent the qualitative representation of Equation \ref{eq:02} outputs. D, DL and DF are the distance between the eCoP of two feet, foot width and foot length respectively. $\theta_{RF}$ and $\theta_{LF}$ are the angle between the foot orientation and the line connecting the eCoPs.}
	\label{fig:01}
\end{figure}

The human legs have an anisotropic shape implies that BoS shape depends not only on the feet position but also from their orientation in the space. The formulation of our model starts from the definition of the Saddle reference frame as shown in Figure \ref{fig:01}.(a). The transformation from the Saddle reference frame (SS)  to the Task-space (TS) coordinates is:

\begin{equation}
\label{eq:01}
\vec{x}_{TS}=R(\theta)\vec{x}_{Saddle}+\left[\begin{array}{l}
\frac{x_{RF}+x_{LF}}{2} \\\\
\frac{y_{RF}+y_{LF}}{2} 
\end{array}\right]
\end{equation}
where $\vec{x}_{TS}$ and $\vec{x}_{Saddle}$ are the coordinates in the TS and the SS reference frames respectively. $x_{RF}$, $y_{RF}$, $x_{LF}$ and $y_{LF}$ are the coordinates of  two eCoPs. Lastly, $\theta$ is the angle between the abscissa axis of the two reference frames.

The transformation in Equation \ref{eq:01} can be used to express the BoS boundaries defined in the SS (formulated in the following equation) in the TS, as exemplified in Figure \ref{fig:01}.(c).

\begin{equation}
\label{eq:02}
\left.\begin{array}[2]{l}
y_{Saddle}=\left\{\begin{array}[2]{ll}
D_{Ly}cos(\theta_{xs}),& \textit{ if }  \theta_{xs} \in [0,\pi)\\
D_{Ry}cos(\theta_{xs}),& \textit{ if } \theta_{xs} \in [\pi,2\pi)\\
\end{array}\right.\\\\
x_{Saddle}=
\left\{\begin{array}[1]{l}
\left\{\begin{array}[2]{ll}
X_L=D_{Ly}sin(\theta_{xs}),& \textit{ if } \theta_{xs} \in [0,\pi) \textit{ \& } \\ & |X_L|\le |D_{Lx} | \\
X_R=D_{Ry}sin(\theta_{xs}),& \textit{ if } \theta_{xs} \in [\pi,2\pi) \textit{ \& }\\ &|X_R|\le |D_{Rx}| \\
\end{array}\right.\\
Otherwise:\\
\left\{\begin{array}[2]{ll}
m_{Bk} y_{Saddle}-D_{Rx}  & \textit{ if } \theta_{xs} \in [\pi/2,3\pi/2)\\
m_{Fw} y_{Saddle}+D_{Rx}  & \textit{ if } \theta_{xs} \in [3\pi/2,\pi/2)\\
\end{array}\right.
\end{array}\right.
\end{array}\right.
\end{equation}
where $\theta_{xs}$ is the angle between a given direction and $x_{Saddle}$; instead, $D_{Ry}$, $D_{Ly}$, $D_{Rx}$, $m_{Bk}$ and $m_{Fw}$ are defined as follows:
\begin{equation}
\label{eq:03}
\left\{ \begin{array}[1]{l}
D_{Ly}=\frac{D}{2}+D_{ly}\\\\
D_{Ry}=-\frac{D}{2}+D_{ry}\\\\
D_{ly}=\frac{D_F sin(\theta_{LF}-\pi/2)+D_L cos(\theta_{LF}-\pi/2)}{2}\\\\
D_{Lx}=\frac{+D_F cos(\theta_{LF}-\pi/2)-D_L sin(\theta_{LF}-\pi/2)}{2}\\\\
D_{ry}=\frac{D_F sin(\theta_{RF}-\pi/2)-D_L cos(\theta_{RF}-\pi/2)}{2}\\\\
D_{Rx}=\frac{D_F cos(\theta_{RF}-\pi/2)-D_L sin(\theta_{RF}-\pi/2)}{2}\\\\
m_{Bk}=\frac{(D_{Rx}-D_{Lx})}{D+D_{Ry}-D_{Ly}}\\\\
m_{Fw}=\frac{(D_{Lx}-D_{Rx})}{D+D_{Ry}-D_{Ly}}
\end{array}\right.
\end{equation}
The geometry of the convex hull defined by each foot is defined based on the measure of the eCoP trajectory during standing as reported in \cite{Hof2005}. The point defined  as eCoP in this model represents the fulcrum of rotation of our inverted pendulum model that is placed at the centre of this area. Although usual inverted pendulum model uses the eCoP as fulcrum of rotation, the proposed approach is based on the growing evidence that human internal model accounts for a fixed reference within the foot \cite{Tiseo2018,Carpentier2017}. 

\subsection*{Validation Method}
The validation of the proposed model involves the analysis of the pelvic movements range of 12 healthy subjects (7 males and 5 females) in 6 different postures (as shown Figure \ref{fig:01}.(b)). The participants' age, mass, and height are 28.3$\pm$3.4 years, 71.1$\pm$12.6 kg and 1.71$\pm$10.7 m respectively. All the subjects provided written informed consent to the experiment.

The subjects were instructed to move their pelvis to explore the borders of the area they were comfortable to move in for about 30 seconds per posture while keeping their torso in an erect position parallel to the frontal plane. They were allowed to partially lift-up the feet as long as the orientations of the feet were kept constant. 

Ten passive markers have been placed on the subjects as described in the \ref{Apx:1}. Their movements have been recorded by an 8-cameras motion capture system from Vicon controlled with Nexus Software. Subsequently, the data have been imported to Matlab (Mathworks Inc) for the data analysis to validate if the SS can be used to track the BoS and, the accuracy of the proposed model based on the geometrical properties of the foot soles. This resulted in the following protocol:
\begin{enumerate}
	\item  The covariances of the CoM trajectories were compared with the ideal BoS predicted by the proposed model for each posture, where the anthropometric parameters required by the model have been set to the mean of subjects' anthropometry.
	\item The Points of Inclusion (POI) is defined as the percentage of recorded CoM positions that are within the boundaries predicted by the model.
	\begin{equation}
	\label{eq:04}
	PoI=  \frac{\textit{N CoM Samples Inside BoS}}{\textit{Total N of CoM Samples}}\times100
	\end{equation}
	\item The Points of Inclusion 360 (PoI360) is the POI considering only the samples on the outer border in the saddle space of the recorded data for each subject. This allows to evaluate if the CoM range of motion is included in the BoS by considering farthest point reached by the CoM trajectories in every direction.
	\begin{equation}
	\label{eq:05}
	PoI360= \frac{\textit{N of CoM Outer Border Samples Inside BoS}}{\textit{N of CoM Outer Border Sample}}\times100
	\end{equation}
\end{enumerate}
\section*{Results}

\label{Sec:3}
\begin{figure}[h!]
	\begin{center}
		
		\includegraphics[width=\textwidth]{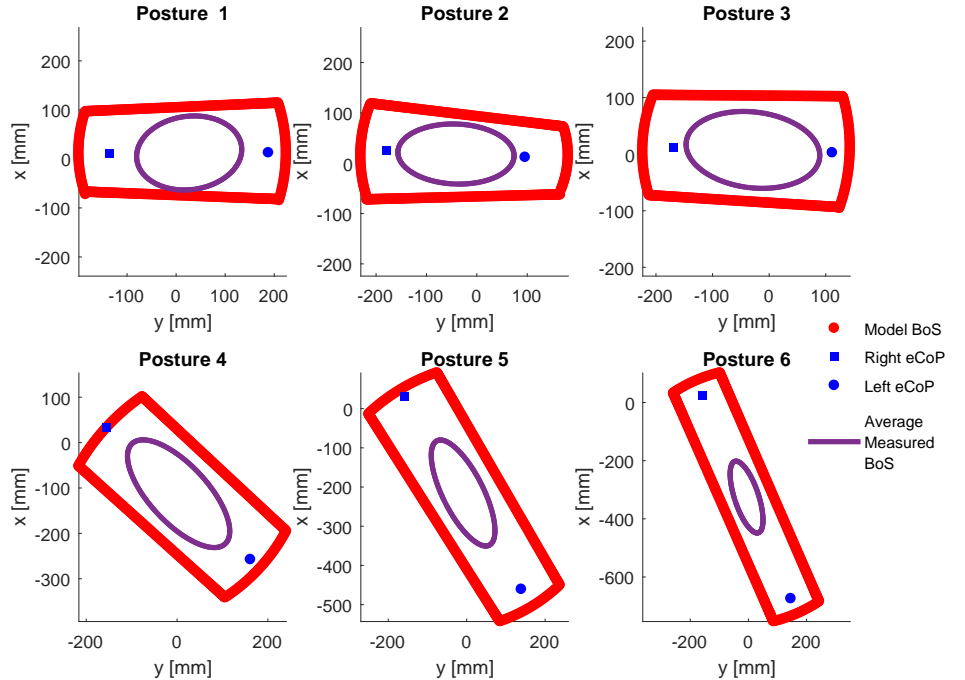}
		
	\end{center}
	\caption{The position of the right and left eCoPs, and the measured and calculated BoS for different postures, computed by averaging the results of the 12 subjects. The average BoS describes the mean of the covariance ellipsis recorded for each of the subjects during the experiment. The data show both the accuracy of the transformation in Equation \ref{eq:01} in tracking the relationship between BoS and postures, as well as the intrinsic limitations of the BoS approach to bipedal stability.}
	\label{fig:04}
\end{figure}

The analysis of the mean areas covered by the subjects in the 6 different postures is shown in Figure \ref{fig:04}. The comparison of the measured data with the proposed model shows that Equation \ref{eq:01} is able to adequately project the shape of the BoS in the Saddle Space (Equation \ref{eq:02}) in the different postures. 

\begin{figure}[h!]
	\begin{center}
		
		\includegraphics[width=\textwidth]{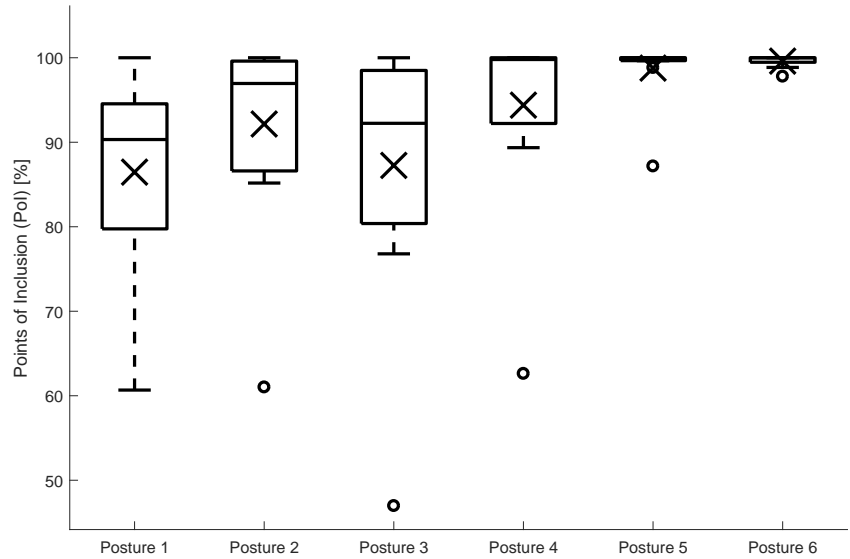}
		
	\end{center}
	\caption{The distribution of PoIs for different postures shows a better accuracy of the geometrical model with the increasing of the postural constraints imposed by the postures.}
	\label{fig:06}
\end{figure}

The values of the mean PoI for postures 1-6 are 86.5$\pm$11.6\%, 92.2$\pm$11.3\%, 87.26$\pm$15.0\%, 94.4$\pm$10.7\%, 98.8$\pm$3.7\% and 99.6$\pm$0.7\% respectively. The data analysis shows that the subjects are able to go beyond the model predicted BoS when they adopt a posture which allows a better range of motion contributing to a lower mean PoI. On the contrary, their PoI increases when they assume postures that introduce higher postural constraints, as shown the Figure \ref{fig:06}.  

\begin{figure}[h!]
	\begin{center}
		
		\includegraphics[width=\textwidth]{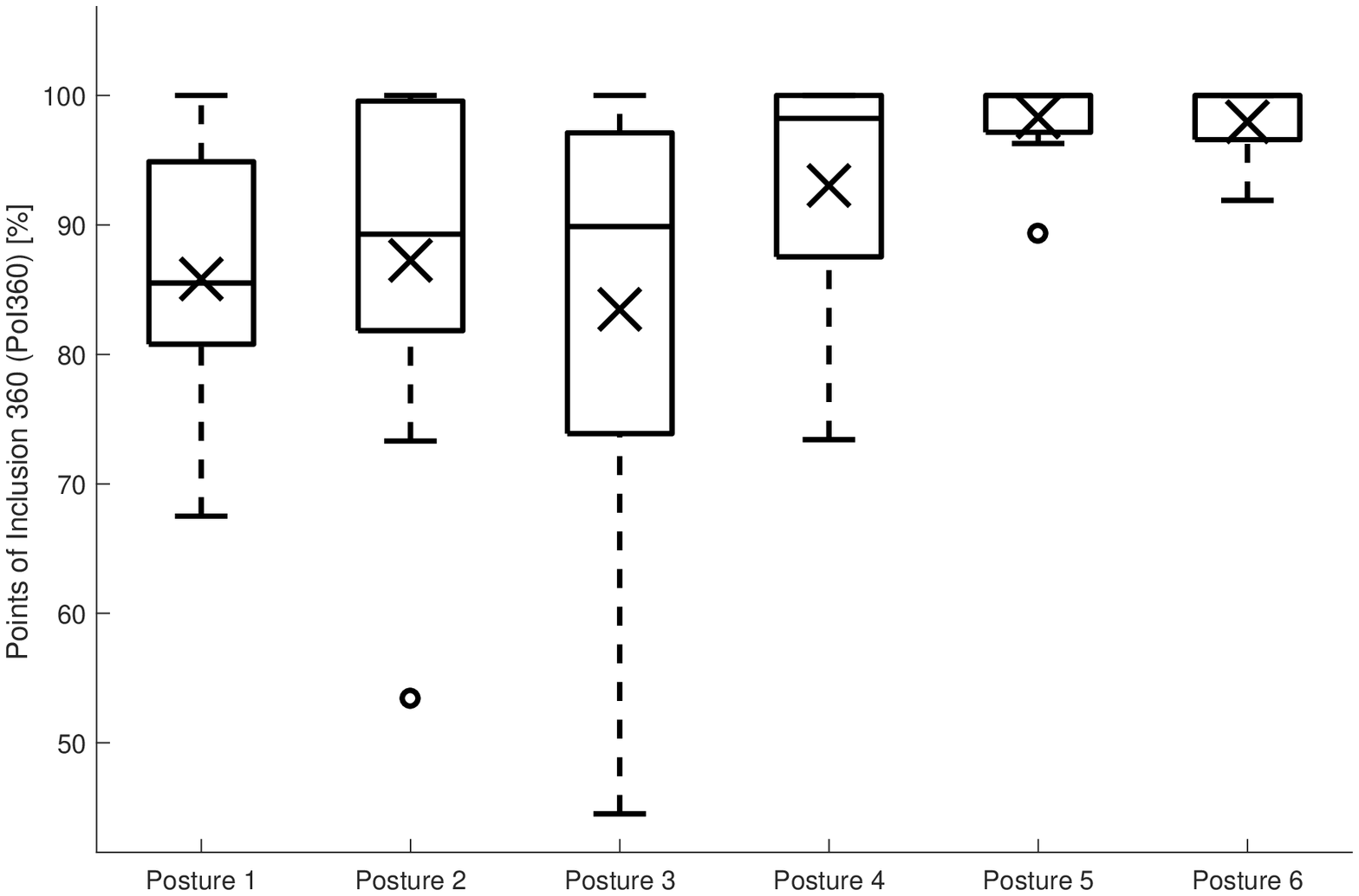}
		
	\end{center}
	\caption{the distribution of PoI360s for different postures shows a relationship with the postural constraint similar to the PoI index shown in Figure \ref{fig:06}.}
	\label{fig:07}
\end{figure}

The results of the PoI are also confirmed by the PoI360, which has mean values of 85.8$\pm$10.6\%, 87.3$\pm$13.8\%, 83.5$\pm$18.0\%, 93.0$\pm$10.27\%, 98.3$\pm$3.2\% and 98.0$\pm$3.2\% for the six postures respectively. Furthermore, the mean PoI360 values increase with the postural constraints of the posture similarly to the PoI, which can be observed by comparing Figures \ref{fig:06} and \ref{fig:07}.

\section*{Discussions}
\label{Sec:4}
Experimental results validate the initial hypothesis that Equation \ref{eq:01} can accurately project geometries defined in the saddle space in the task space. Nevertheless, the current geometrical models of the BoS still retains most of the limitations of the previous formulations mentioned in Section \ref{Sec:1}. The proposed can be further improve accounting for the alteration of its geometry due to the changes of the inclination of the feet' soles with respect to the ground.

The direct relationship of both PoIs to the increase in postural constraints highlights the limitation of the BoS geometrical models in the stability evaluation for highly redundant systems. The BoS concept accurately describes the stability of rigid bodies (i.e. seesaw, rigid inverted pendulum), where the Centre of Mass (CoM) is fixed within the geometry, and the gravitational forces are counteracted by the structure rigidity \cite{Pollock2000}. On the other hand, the CoM confinement within the BoS is a necessary but not a sufficient condition for stability in non-rigid structures, where such analysis has also to account for the forces required to retain the desired configuration.

The CoM location is not constant in reconfigurable structures, and its calculation requires the complete knowledge of the kinematics and dynamics of the systems. Therefore, the CoM tracking becomes extremely complex for high-dimensional structures such as human body. A common way to deal with such problem is to sacrifice some accuracy by using a fixed location of a landmark-based CoM defined considering anatomical landmarks and ergonomics, as the one used in this paper  \cite{Hof2005,Lugade2011,McAndrewYoung2012,Hak2013,Hak2013a,Torricelli2016,Font-Llagunes2009,VanMeulen2016}.

The stability conditions analysed in the previous two paragraphs only considers the static stability, which resides in the ability to retain a posture. However, there is also a dynamic stability condition, which requires not only the ability to counteract the external forces acting on the systems but also the management its own kinetic energy. There are different ways to account for the system energetics, but the most commonly used impose kinematics constraints based on the inverted pendulum analogy\cite{Hof2005,Kuo2007,VanMeulen2016}, and they require input of the ground reaction forces that are not easily collected in unstructured environments. On the other hand, the proposed model provides an alternative framework for the implementation of available methods to evaluate stability. For example, the eXtrapolated-CoM (XCoM) can be implemented with IMUs and optoelectronic system (i.e. motion capture) using the proposed method to derive the required parameters (BoS, CoM position and Velocity)  from the body kinematics. 

Furthermore, equation \ref{eq:01} allows the transformation of physical quantities from a posture dependent SS to the TS. Therefore, it opens with a new framework to describe posture dependent parameters used both for humans and robots. For example, as the proposed model allows to project the BoS to the TS accurately, it can be used to model capture region generated by the two legs, which describes the area of space where a bipedal robot can attract the CoM towards a fixed point under specific dynamic conditions \cite{Pratt2006}. 

\section*{Conclusions}
\label{Sec:5}

The experimental results demonstrate that the model presented in this paper can be instrumental in the characterisation of bipedal systems in general. Furthermore, our results show that it has a good level of accuracy, despite the drawbacks caused by having the model relying on geometrical definition for the BoS and the CoM. 

\begin{figure}[h!]
	\begin{center}
		
		\includegraphics[width=0.8\textwidth]{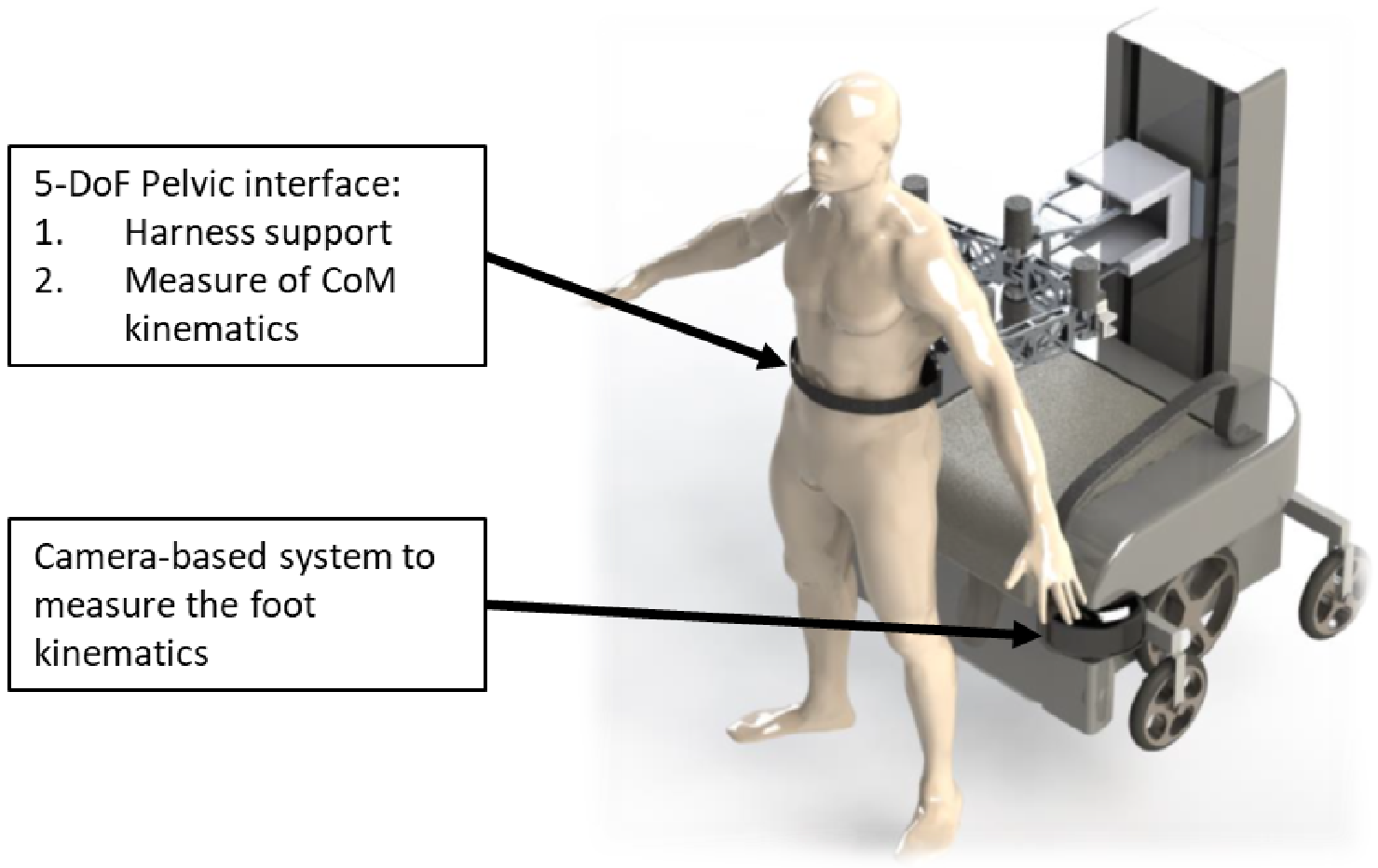}
		
	\end{center}
	\caption{Concept design of the target application of the proposed method, which will be use to asses the stability of the movement and determine the behaviour of the follow-me wheelchair (e.g., follow the movement or prepare for intervention).}
	\label{fig:08a}
\end{figure}

Nonetheless, at the current stage, these models can be only employed for rehabilitation robots when harness support is provided due to safety constraints. Therefore, it satisfies the requirement for our application where the robot should provide harness support without interfering with the patients’ locomotor strategies through a sensorized pelvic interface \cite{Tiseo2014,Tiseo2015}. Furthermore, we are currently implementing this method in a camera-based sensor system to measure the feet postures that will be mounted on the robotic base, while the position of the CoM will be measured via the pelvic interface kinematics, as shown in Figure \ref{fig:08a}. 

Lastly, the proposed Saddle Space has been proven to be a convenient way to describe, observe and measure bipedal systems stability and it can be used regardless of the geometry defined for the BoS of the two feet. Thus, the development of a dynamic model in such space may ease and improve the control, planning and performance assessment of bipedal locomotion in the future.

\section*{Acknowledgements}{This research was supported by the A*STAR-NHG-NTU Rehabilitation Research Grant: "Mobile Robotic Assistive Balance Trainer" (RRG/16018). This experiment has been approved by the NTU Internal Review Board, Reference Number: IRB-2016-09-015. Figure \ref{fig:6New}.(a) has been reprinted from Journal of Biomechanics, 38, A.L. Hof M.G.J. Gazendam and W.E. Sinke, The condition for dynamic stability, 1-8, 2005 \cite{Hof2005} with permission from Elsevier.}

\section*{Conflicts of Interest}{The authors declare no conflict of interest.} 

\section*{Abbreviations}{The following abbreviations are used in this manuscript:\\
	\noindent 
	\begin{tabular}{@{}ll}
		BoS & Base of Support\\
		CoM & Centre of Mass\\
		CoP & Centre of Pressure\\
		eCoP & extrapolated Centre of Pressure\\
		TS & Task Space\\
		SS & Saddle Space\\
		PoI & Points of Inclusion\\
		PoI360 & Points of Inclusion 360\\
\end{tabular}}

\appendix
\section{Marker Set and Model Reference Points Calculation}
\label{Apx:1}
Ten IR reflective markers were placed on the subjects to capture the CoM position, feet positions and orientation. The proposed model uses the eCoP position and the base of support of the feet to derive the BoS from the feet's posture. The eCoP and the feet's BoS geometry have been defined based on the measure of the CoP range of motion (Figure \ref{fig:6New}.(a)) reported by Hof et al in \cite{Hof2005} as shown in Figure \ref{fig:6New}.(b). Figures \ref{fig:6New} and \ref{fig:6New} also shows the positions of the markers placed on the pelvis and feet of the subjects, with the description given below that is a subset of the marker set used in \cite{Mandery2015}:

\begin{figure}[h]
	\begin{center}
		\includegraphics[width=\textwidth]{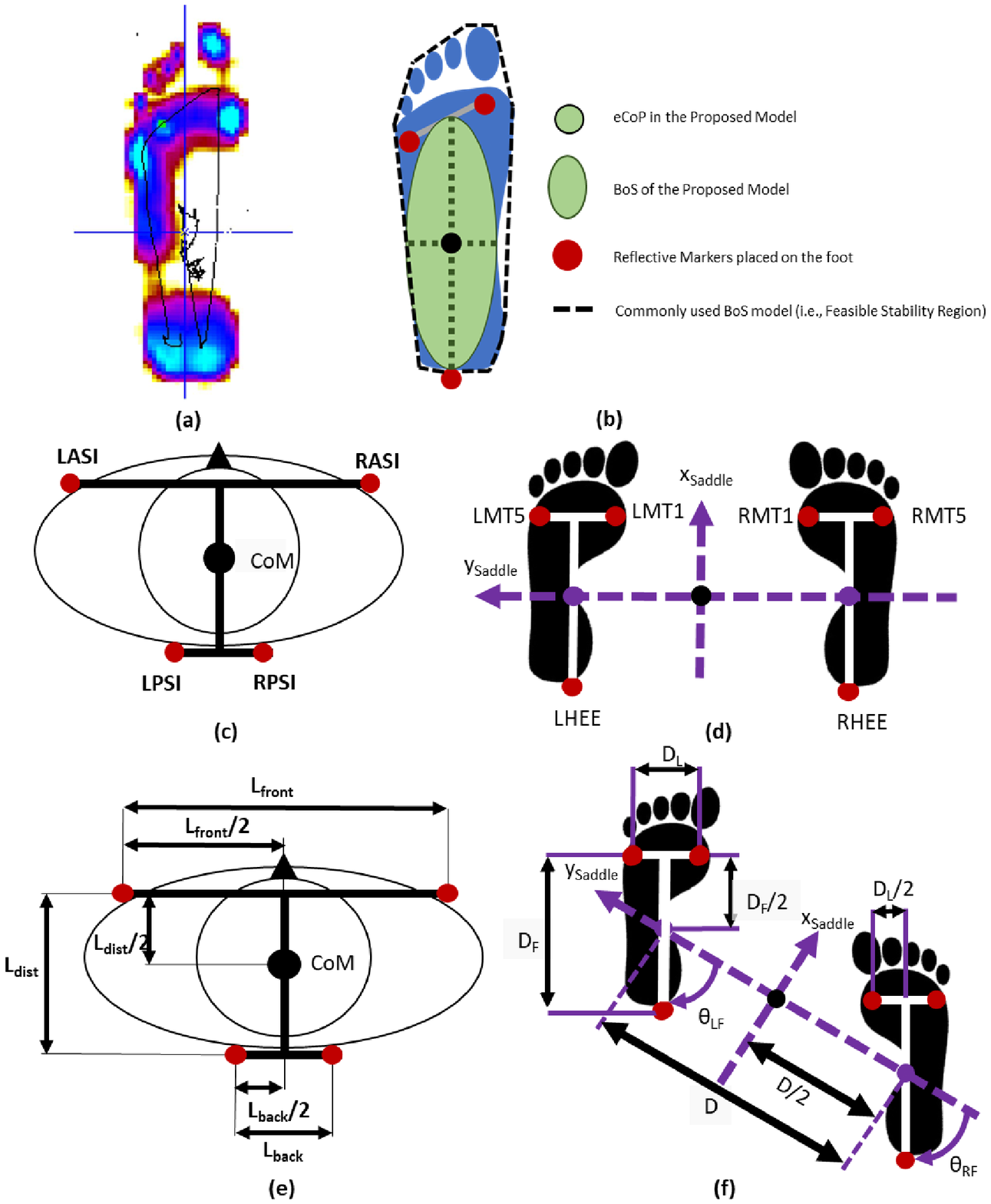}
	\end{center}
	\caption{(a) The physical CoP trajectory in black that has been recorded with a RSscan$^{TM}$ Footscan, and it has been used to define the BoS in \cite{Hof2005}. (b) The eCoP and the BoS in the proposed model have been determined based on the CoP trajectory reported in \cite{Hof2005}. Furthermore, the proposed model is a better approximation of the measured BoS when compared with the commonly used Feasibility Stability Region \cite{Yang2009}. (c) Hip Marker-set, (d) Feet Marker-set. (e) Calculation of the CoM position from the 4 pelvic markers A6.(f)Calculation of the feet parameters from marker positions.}
	\label{fig:6New}
\end{figure}

\begin{itemize}
	\item LASI: Directly above spina iliaca anterior superior (left)
	\item RASI: Directly above spina iliaca anterior superior (right)
	\item LPSI: Directly above spina iliaca posterior superior (left)
	\item RPSI: Directly above spina iliaca posterior superior (right)
	\item LHEE: Center of left heel
	\item RHEE: Center of right heel
	\item LMT1: Left first metatarsal bone
	\item LMT5: Left fifth metatarsal bone
	\item RMT1: Right first metatarsal bone
	\item RMT5: Right fifth metatarsal bone
\end{itemize}

The CoM and both the eCoPs for the feet are calculated from the marker positions as shown in Figure \ref{fig:6New}. The foot width $D_L$ is the distance between the two markers on the metatarsals, whereas foot length $D_F$ is the distance between the marker on the heel and the midpoint of $D_L$. The distance $D$ between the feet is the distance between the midpoints of the two $D_L$. The foot orientation is the angle between $D$ and $D_L$ ($\theta_{LF}$ and $\theta_{RF}$).

\bibliography{library}

\bibliographystyle{abbrv}

\end{document}